\documentclass{article}

\usepackage[utf8]{inputenc}
\usepackage[T1]{fontenc}
\usepackage{hyperref}
\usepackage{url}
\usepackage{amsfonts}
\usepackage{nicefrac}
\usepackage{microtype}
\usepackage{graphicx}
\usepackage{float}
\usepackage[justification=centering]{caption}
\usepackage[numbers]{natbib}
\usepackage{todonotes}
\usepackage{booktabs}

\title{Detecting Racial Bias in Jury Selection}

\author{Jack Dunn, Ying Daisy Zhuo\\
        Interpretable AI\\
        Cambridge, MA 02142\\
        \texttt{info@interpretable.ai} \\
}

\date{}

\begin{document}

\maketitle

\begin{abstract}
To support the 2019 U.S. Supreme Court case ``Flowers v. Mississippi'', APM Reports collated historical court records to assess whether the State exhibited a racial bias in striking potential jurors. This analysis used backward stepwise logistic regression to conclude that race was a significant factor, however this method for selecting relevant features is only a heuristic, and additionally cannot consider interactions between features. We apply Optimal Feature Selection to identify the globally-optimal subset of features and affirm that there is significant evidence of racial bias in the strike decisions. We also use Optimal Classification Trees to segment the juror population subgroups with similar characteristics and probability of being struck, and find that three of these subgroups exhibit significant racial disparity in strike rate, pinpointing specific areas of bias in the dataset.

\end{abstract}

\section{Introduction}

It is no secret that human decision making is fraught with biases, both conscious and subconscious. It is less well-known that automated systems based on machine learning and artificial intelligence are susceptible to the same issues, although this is becoming more evident with time. For instance, it has been demonstrated that commercial systems for facial recognition exhibit significantly higher error rates for minorities, and that a likely cause was unrepresentative data samples used for training~\cite{buolamwini2018gender}. Another study found that a commercial algorithm used to guide healthcare decisions prioritized the needs of white patients over others, again due to training the model using a biased dataset~\cite{obermeyer2019dissecting}.

One of the key concerns with the increasing prevalance of such algorithms in daily life is that rather than enabling fair and data-driven decision making, they may simply learn from the existing biases in historical data, and cement these in place. To reduce and avoid these issues, it is increasingly important that we have the ability to inspect machine learning models to comprehend and interpret their decision making, and thus surface any problematic biases that may be present. The increased activity in this area has resulted in a number of transparent and interpretable machine learning approaches that can be used to assess a dataset proactively for problematic biases.

In this paper, we consider the 2019 U.S. Supreme Court case ``Flowers v. Mississippi'' which dealt with the question of whether jury strike decisions were racially motivated. In particular, we focus on one piece of supporting evidence in the case: historical court records that were collated and analyzed for evidence of racial bias, the findings of which were submitted in support of the case.

\subsection*{Overview of ``Flowers v. Mississippi''}

In 1986, the U.S. Supreme Court ruled in the landmark case ``Batson v. Kentucky'' that any use of peremptory challenges to exclude jurors based solely on race reason to remove potential jurors is unconstitutional~\cite{batson}. Despite this ruling, a large disparity in juror strike rates across races appears to remain in many courts across the U.S. For instance, in two counties in Florida, 64\% of felony cases between 2000 and 2010 had at least one black potential juror, but only 28\% of trials had a black juror seated, indicating a disproportionate propensity for striking black jurors~\cite{anwar2012impact}.

This disparity was the focus of the 2019 U.S. Supreme Court case ``Flowers v. Mississippi'', where it was ruled that the District Attorney Doug Evans from the Fifth Circuit Court District in Mississippi had discriminated based on race during jury selection in the six trials of Curtis Flowers~\cite{flowers}. Flowers was accused of four murders that were committed in 1996 in Winona, Mississippi, and was tried six separate times for these crimes. In the first three trials, Flowers was convicted and sentenced to death, but in each case the conviction was subsequently overturned by the Mississippi Supreme Court for reasons of prosecutorial misconduct, and the case was later retried. The fourth and fifth trials ended in mistrials due to hung juries: in particular the jury of the fourth trial was split 7--5 in favor of conviction, and this split was along racial lines with the five black jurors voting to acquit. The sixth trial again resulted in a conviction and sentencing to death. 

Flowers appealed the conviction of the sixth case to the Mississippi Supreme Court, citing racial prejudice in jury selection. In particular, it was argued that the State had accepted white jurors that were almost identical to black jurors rejected by the State, and as such, the only explanation was that the State had struck the black jurors based solely on their race. The Mississippi Supreme Court ruled against Flowers in 2014, but was later ordered in 2016 by the U.S. Supreme Court to reconsider its decision after the 2016 case ``Foster v. Chatman''. This was a different case altogether which had ruled in favor of the defendant, where it was decided that the prosecution had intentionally discriminated based on race in the selection of jurors~\cite{foster}. Upon review, the Mississippi Supreme Court in 2017 affirmed its initial decision that no discrimination had taken place, and Flowers subsequently appealed the case to the U.S. Supreme Court.

To support the appeal to the U.S. Supreme Court, journalists from APM Reports collected and published court records of jury strikes in the Fifth Circuit Court District of Mississippi and analysed them to assess if there was a systematic racial bias in jury selection in this district~\cite{apmreports}. The data includes information on each trial, juror, and the voir dire answers by the jurors between 1992 and 2017. As part of their analysis, they used a logistic regression to predict the probability of a juror being struck. From the coefficient assigned to race by the model, they concluded that there was significant racial disparity in jury strike rates by the State, even after accounting for other factors in the dataset~\cite{apmwhitepaper}.

The U.S. Supreme Court in ``Flowers v. Mississippi'' ruled in favor of Flowers and overturned the conviction, finding that State's use of peremptory strikes was indeed motivated in substantial part by race. In particular, the Supreme Court opinion cited four pieces of evidence supporting their ruling:

\begin{enumerate}
    \item Over the six trials, the State struck 41 out of 42 black jurors that it could possibly have struck.
    \item In the sixth trial, five of six potential black jurors were struck by the State.
    \item In the sixth trial, the number of questions asked of jurors by the State was dramatically different across black and non-black jurors, in an apparent attempt by the State to find reasons to strike the black jurors.
    \item One of the black jurors that was struck was almost indistinguishable from white jurors that were seated, based on the questions asked by the State, and thus the decision was racially motivated.
\end{enumerate}

The dissenting opinion of Justice Thomas argued that the prosecution's use of peremptory strikes in this case was appropriate, as most of those dismissed knew Flowers or his family and would be biased if included in the jury.

\subsection*{Overview of the APM Reports analysis}

The analysis conducted by APM Reports to support the appeal to the U.S. Supreme Court was based on backward stepwise logistic regression~\cite{apmwhitepaper}. Their stated procedure for constructing the model was to first run the model with all features present, and remove those that had a p-value above 0.1. They then re-ran the model on the new subset of variables, and removed those with p-value above 0.05. This left seven factors as being predictive of being struck:

\begin{itemize}
    \item whether the prospective juror is black
    \item whether the prospective juror has ever been accused of a crime
    \item whether a family member of the prospective juror has ever been accused of a crime
    \item whether a family member of the prospective juror works as law enforcement
    \item whether the prospective juror has any hesitation about applying the death penalty
    \item whether the prospective juror admits to knowing the defendant
    \item whether the prospective juror is the same race as the defendant
\end{itemize}

One limitation of this approach is that this method of selecting which features are included in the model is subjective and a question of human judgement. Different heuristic approaches to variable selection can lead to different results that may affect the conclusions of the analysis, changing which factors are deemed predictive of being struck. A data-driven, optimal approach to feature selection would not have these issues.

Another limitation is that the logistic regression model used is inherently linear and additive by nature. This means that it cannot discover interaction effects between variables. In particular, it cannot discover if there are subpopulations in the data that exhibit different racial disparities in strike rate. Indeed, the APM Reports analysis states that they ``found no way to slice up the data where the State struck white and black jurors in equal measure.'' Applying a non-linear model to the dataset could enable better understanding and characterization of how the racial disparity in strike rate might vary across the juror population.

\subsection*{Our approach}

In this paper, we aim to augment the analysis conducted by APM Reports and address the limitations. In particular, we seek to answer the following questions:
\begin{itemize}
    \item Using a feature selection approach that is globally optimal rather than a heuristic, do we reach the same conclusion that race has significant power in predicting the strike rate of jurors?
    \item By applying a non-linear method, can we discover specific subgroups that exhibit different racial disparity in strike rate?
\end{itemize}

To do this, we leverage recent advances in the machine learning literature that generate highly-performant and interpretable models through the use of modern optimization. The interpretability of these methods allows us to assess the degree of bias in the dataset in a transparent and understandable way, helping us to comprehend the patterns in the data in an intuitive manner.

Our contributions in this paper are as follows:
\begin{enumerate}
    \item Using Optimal Feature Selection, we identify that the optimal set of features to include in a logistic regression model to predict strike race does indeed include race, affirming the conclusion of APM Reports. Moreover, we find that at any level of feature inclusion, race is always selected as one of the most important factors in predicting outcomes, providing strong evidence that racial bias is indeed present. Finally, when we explicitly remove the race variable from the model, the performance (in terms of AUC) drops from 0.81 to 0.67, demonstrating that the race of the juror is providing unique signal in the data that cannot be proxied for by the remaining variables.
    
    \item Using Optimal Classification Trees, we segment the juror population in a data-driven fashion into five subgroups with similar propensity to being struck. Two of these subgroups exhibit no racial disparity in strike rate, while the remaining three subgroups have statistically significant differences in strike rate between black and non-black jurors. The interpretability of this model provides insights into the decision process and biases of the State. One of these groups with racial disparity is comprised of jurors that have never been accused of a crime, but know the defendant. In this group, 85\% of black jurors were struck by the State, whereas only 20\% of non-black jurors were struck by the State. This finding is particularly significant given the dissent of Justice Thomas, where he argued that the State was justified in striking the prospective black jurors that knew the defendant to avoid bias on the jury. Our evidence shows that this reasoning was not applied equally to jurors of all races.
\end{enumerate}

The structure of this paper is as follows. In Section~\ref{ofs}, we follow the approach of APM reports and apply Optimal Feature Selection to predict probability of jurors being struck to identify in an optimal manner the most predictive features. In Section~\ref{oct}, we apply the non-linear method Optimal Classification Trees to identify subgroups of the juror population that exhibit different racial disparities in strike rate. Finally, in Section~\ref{conclusions}, we present a summary of our findings.

\subsection*{Literature Review}

The rise of machine learning to assist or replace human decision makers has prompted significant research aimed at addressing concerns of fairness and bias in automated decision making (see~\cite{mehrabi2019survey} for a survey). One of the primary concerns is that when attempting to learn from past decisions, machine learning models may simply learn and perpetuate existing biases present in the historical decision making process. As such, it is critical to have the ability to inspect models after training to understand if they are indeed making decisions in ways that are deemed fair. Being able to interpret and comprehend the logic behind model predictions thus enables us to assess and understand the degree of pre-existing bias in a dataset.

Linear models such as linear and logistic regression are easily understood and interpreted. However, they can suffer from interpretability issues when too many features are selected in the model. There are many traditional approaches to filtering the included variables down to a relevant subset. Forward or backward stepwise methods are heuristics that are used to construct the model iteratively, often in conjunction with human intuition~\cite{hastie2009elements}. This is the approach taken by APM Reports. Regularization-based methods such as lasso~\cite{tibshirani1996regression} or elastic net~\cite{zou2005regularization} are data-driven heuristic approaches to reducing the number of features included in the model, which they aim to achieve by penalizing the regression coefficients. However, these approaches do not directly address the feature selection problem, and in fact construct solutions that are robust to data perturbations, a desirable but orthogonal concern~\cite{bertsimas2018characterization}. Optimal Feature Selection~\cite{bertsimas2016best,bertsimas2019sparse,bertsimas2020sparse,bertsimas2017sparse} is recent family of approaches that formulates the best subset selection problem using integer optimization, enabling the exact solution of the problem to provable optimality and a guarantee that the best subset of features has been identified. 

In addition to linear models, there are also a variety of non-linear models that can easily model interactions between features in the dataset. Traditional examples of these methods include interpretable decision tree approaches such as CART~\cite{breiman1984classification} as well as black-box methods like random forests and gradient boosting~\cite{hastie2009elements} that are not easily understood by humans. The classifical interpretable methods like linear regression and decision trees are usually outperformed by black-box methods in terms of predictive performance, forcing practitioners to consider a trade-off between interpretability and performance. However, another recent stream of research into constructing decision trees with global optimization has led to Optimal Classification Trees~\cite{bertsimas2017optimal,bertsimas2019machine} which can create interpretable decision trees like those of CART, but with performance that rivals black-box methods, obviating the need to trade-off between performance and interpretability.

\section{Detecting Presence of Racial Bias using Optimal Feature Selection}\label{ofs}

In this section, we follow a similar approach to that used in the APM Reports analysis, but instead of using a backward stepwise method to select features for inclusion in the model, we use Optimal Feature Selection to select the optimal subset of features.

To ensure consistency, we followed the same data processing methodology as in~\cite{apmwhitepaper}. For each juror, we assembled the trial and juror information as well as the juror's voir dire answers. We subset the data to only those jurors eligible to be struck by the State. The prediction target for the model is whether the juror was struck by the State or not.

Unlike the APM Reports study, we split the data into training and testing sets, containing 70\% and 30\% of the data, respectively. This allows us to train models on the training set and evaluate the model on the testing set to gain a fair and robust estimate of the out-of-sample performance of the model. This process helps to avoid overfitting models to the data.

\begin{figure}
  \centering
  \includegraphics[width=\textwidth]{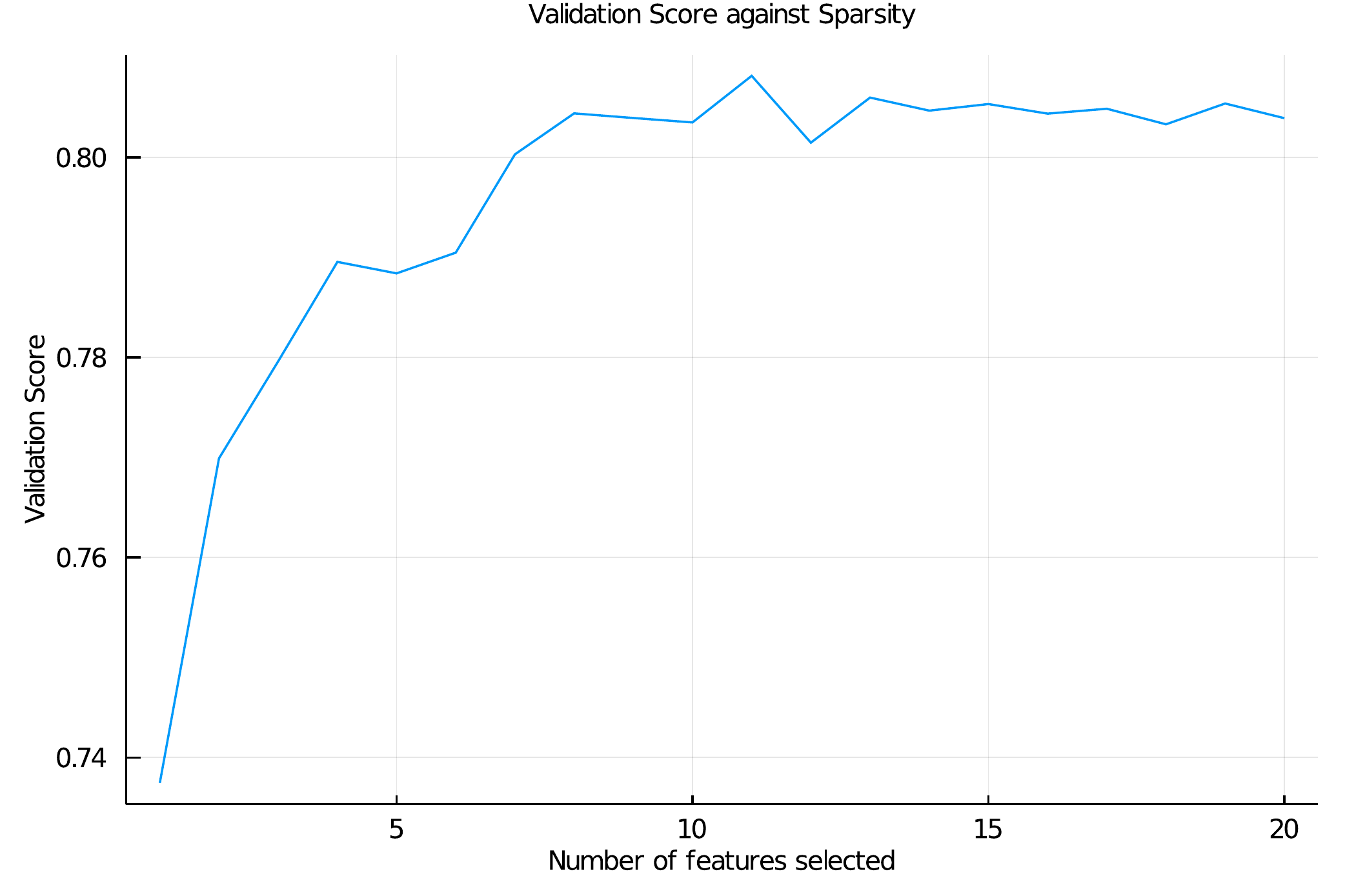}
  \caption{AUC of Optimal Feature Selection model in cross-validation against the number of features selected.}
  \label{fig:ofs-validation}
\end{figure}

We applied Optimal Feature Selection to build logistic regression models based on optimal subsets of features in the dataset. We considered feature subsets of sizes from 1--20, and used 5-fold cross validation to choose the best number of features. Figure~\ref{fig:ofs-validation} shows the area under the ROC curve (AUC) of the models against the number of features selected. We see that the validation performance initially increases with the number of features selected, before peaking at 11 features selected. Additional features past this point do not improve the performance of the model. The AUC of this best model on the testing set was 0.815, indicating strong out-of-sample performance.

\begin{table}
    \centering
    \begin{tabular}{lr}
        \toprule
        Feature & Coefficient\\
        \midrule
        \texttt{accused} & 5.75082\\
        \texttt{is\_black} & 1.25867 \\
        \texttt{fam\_accused} & 1.46095 \\
        \texttt{know\_def} & 1.49665 \\
        \texttt{medical} & 4.34570\\
        \texttt{no\_death} & 8.45462 \\
        \texttt{death\_hesitation} & 3.23671 \\
        \texttt{leans\_defense} & 3.85258 \\
        \texttt{same\_race} & 0.31295\\
        \texttt{fam\_law\_enforcement} & -0.36025 \\
        \texttt{no\_responses} & -0.15774 \\
        \texttt{intercept} & -0.81973\\
        \bottomrule
    \end{tabular}
    \caption{Coefficients for the best logistic regression model from Optimal Feature Selection, listed in order of relative importance. A positive coefficient indicates the probability of being struck becomes higher if a prospective juror exhibits this feature.}\label{ofs-model}
\end{table}

Table~\ref{ofs-model} shows the features and corresponding coefficients for the Optimal Feature Selection model with 11 features, in order of relative importance to the model. All seven features selected by APM Reports show up in our model:

\begin{itemize}
    \item \texttt{accused}: whether the prospective juror has ever been accused of a crime
    \item \texttt{is\_black}: whether the prospective juror is black
    \item \texttt{fam\_accused}: whether a family member of the prospective juror has ever been accused of a crime
    \item \texttt{know\_def}: whether the prospective juror admits to knowing the defendant
    \item \texttt{death\_hesitation}: whether the prospective juror has any hesitation about applying the death penalty
    \item \texttt{same\_race}: whether the prospective juror is the same race as the defendant
    \item \texttt{fam\_law\_enforcement}: whether the prospective juror has any family in law enforcement
\end{itemize}

Our model also selected the following four features as significant that were not selected by APM Reports:
\begin{itemize}
    \item \texttt{medical}: the prospective juror had medical conditions that would make serving difficult
    \item \texttt{no\_death}: the prospective juror said they could not or would not impose the death penalty
    \item \texttt{leans\_defense}: the prospective juror expressed a bias in favor of the defense
    \item \texttt{no\_responses}: the prospective juror did not respond to voir dire questions
\end{itemize}

By selecting these four additional features for inclusion, the Optimal Feature Selection method indicates that they contribute significant signal to the problem and increase the predictive power. Indeed, it seems that these additional features form a more complete model for whether a juror is struck. The most significant of the additional variables is \texttt{medical}, and it is intuitive that a juror indicating they cannot serve on the juror for medical reasons would lead to them being struck. Similarly, it is hard to justify a model that includes \texttt{death\_hesitation} without also including \texttt{no\_death}, since refusing to impose the death penalty is a stronger signal than expressing hesitation. It also makes sense that expressing a bias towards the defense or a lack of responses to questioning would affect the chance of being struck.

\begin{figure}
  \centering
  \includegraphics[width=\textwidth]{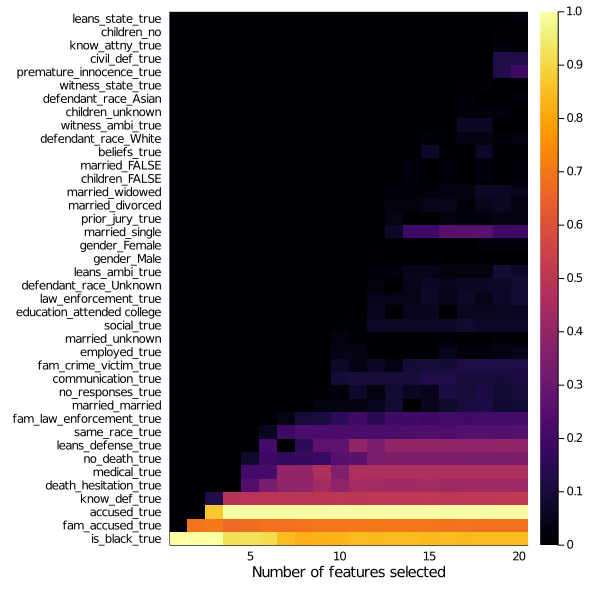}
  \caption{Relative importance of features against the number of features selected.}
  \label{fig:variable-importance}
\end{figure}

To better understand the relative importance of features, Figure~\ref{fig:variable-importance} shows the relative importance of the features as the number of features selected by the Optimal Feature Selection changes. We see that the importance of features is relatively consistent as the number of features is increased. \texttt{is\_black} and \texttt{accused} are the two most importance features, followed by \texttt{fam\_accused}. We see that the results roughly stabilize at about the tenth feature, indicating that the incremental features added are not bringing additional signal to the model, reinforcing the trend seen in Figure~\ref{fig:ofs-validation}.

We note that the model selecting \texttt{is\_black} as an important feature is not alone evidence of bias in the strike decisions. It could be the case that \texttt{is\_black} is correlated with other features that are influencing the strike decisions, and thus might simply be being used by the model as a proxy for the true underlying features. To investigate this, we removed the features in the dataset that related to the juror's race (\texttt{is\_black} and \texttt{same\_race}) and then reapplied Optimal Feature Selection. In doing so, the out-of-sample AUC of the model fell from 0.815 to 0.672. This sharp fall in predictive performance is evidence that the race of the juror is providing unique signal in the dataset that is not captured by other features, reinforcing the conclusion that race is a significant factor in the decision making of the State.

In summary, by applying Optimal Feature Selection to identify the optimal subset of features for the logistic regression model, we confirm the results of the APM Reports analysis that race is a significant factor in predicting the strike rate. Race was one of the most important features regardless of the number of features selected, and moreover the performance falls significantly if the race of the juror is removed. Finally, the set of features chosen by Optimal Feature Selection seems more intuitive than the original analysis, as it also includes key juror features that we would expect to affect the decision process, like medical considerations and refusal to impose the death penalty.

\section{Identifying Subgroups with Significant Racial Disparity using Optimal Classification Trees} \label{oct}

Based on the results of the Optimal Feature Selection, we have strong evidence that the race of the juror plays a significant role overall in predicting the probability of being struck by the State. Next, we would like to investigate if there are specific subpopulations where this effect is more or less pronounced. In this section, we apply Optimal Classification Trees to segment the data into groups with similar propensity to being struck, and then assess these groups for racial disparities in strike rate.

As stated earlier, the logistic regression models of APM Reports and Optimal Feature Selection are linear and additive, meaning the impact of each feature in the model is the same across the entire juror population. In order to understand if race has different importance for specific subgroups in the population, we need to apply a model that is non-linear and can consider interactions between features.

\begin{figure}
  \centering
  \includegraphics[width=0.7\textwidth]{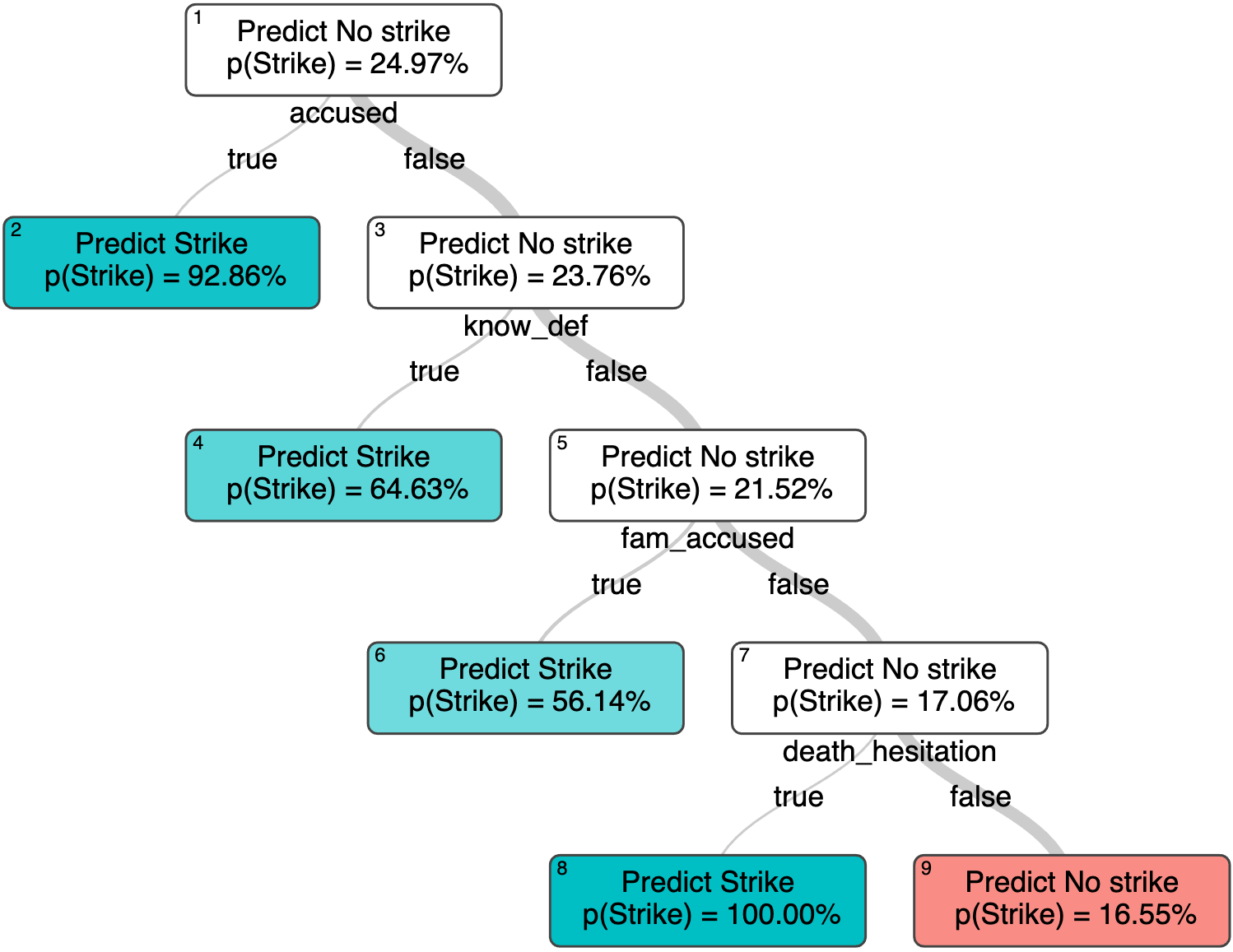}
  \caption{Optimal Classification Tree for predicting probability of strike.}
  \label{fig:tree}
\end{figure}

We used Optimal Classification Trees (OCT) to discover subgroups in the population with similar probabilities of being struck. OCT is a non-linear and interpretable method that produces a single decision tree, thus the subgroups that it discovers are defined based on simple combinations of the features in the dataset. We used the same 70\% of the data as the training set and trained an OCT to predict the probability of a juror being struck. We did not include the features related to the juror's race in this model, as these will be used to assess the racial disparity in each subpopulation. The resulting OCT model is shown in Figure~\ref{fig:tree}.

We see that the OCT identifies five distinct subgroups in the data, where the jurors falling into each group have similar probabilities of being struck:
\begin{itemize}
    \item \textbf{Node 2}: potential jurors that have been accused of a crime have a 93\% strike rate.
    \item \textbf{Node 4}: potential jurors that have never been accused of a crime but know the defendant have a 65\% strike rate.
    \item \textbf{Node 6}: potential jurors that have never been accused of a crime and do not know the defendant, but do have a family member that has been accused of a crime have a 56\% strike rate.
    \item \textbf{Node 8}: potential jurors that have never been accused of a crime, do not know the defendant, do not have a family member that has been accused of a crime, but express hesitation about imposing the death penalty have a 100\% strike rate.
    \item \textbf{Node 9}: potential jurors that have never been accused of a crime, do not know the defendant, do not have a family member that has been accused of a crime, and express no hesitation about imposing the death penalty have a 17\% strike rate.
\end{itemize}

The paths through the tree leading to strike predictions intuitively make sense as reasons to exclude potential jurors. As shown by the Optimal Feature Selection, having been accused of a crime, knowing the defendant, having a family member that has been accused of a crime, or expressing hesitation about imposing the death penalty are all factors that generally increase the risk of being struck. 

\begin{figure}
  \centering
  \includegraphics[width=\textwidth]{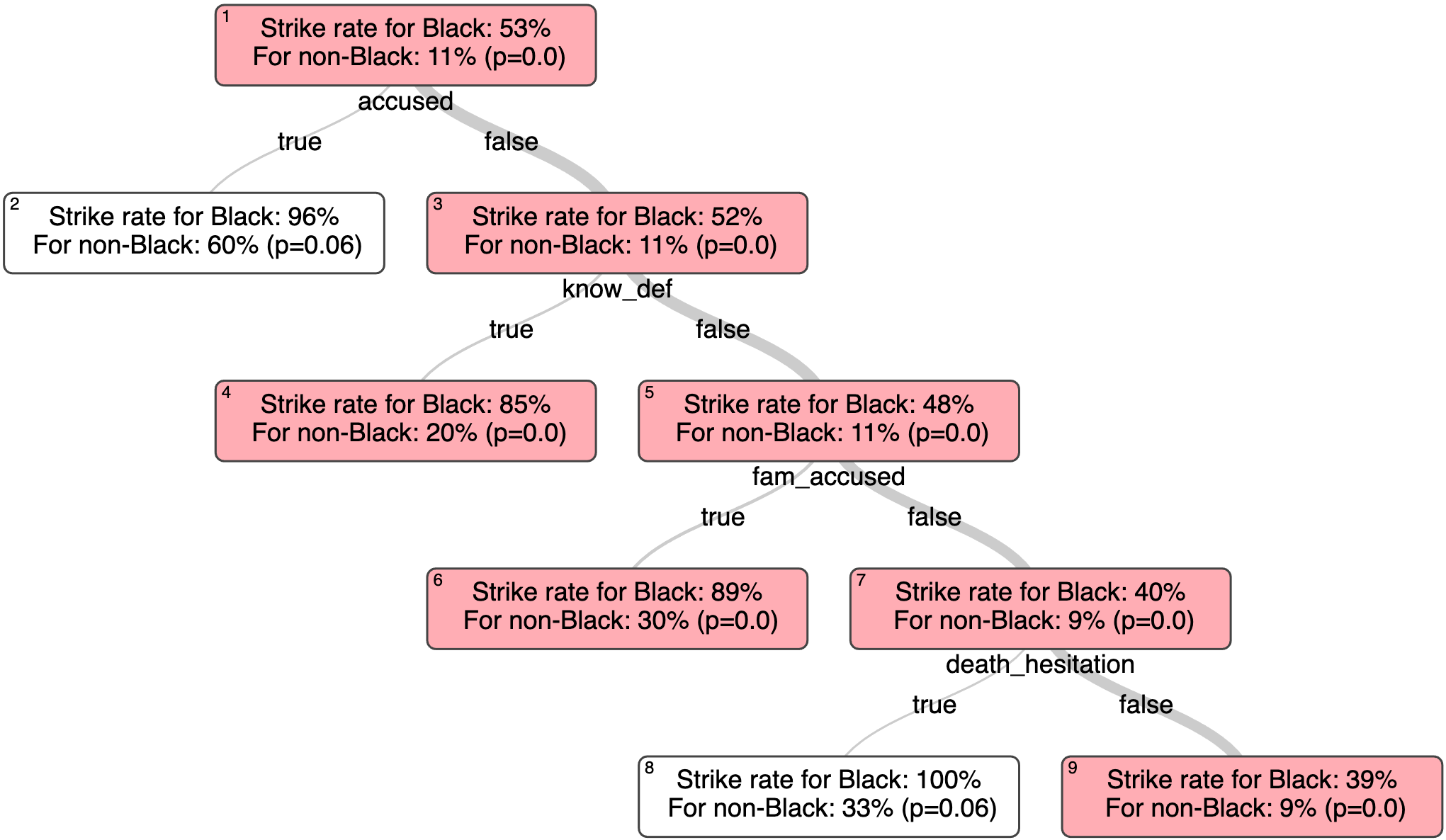}
  \caption{Optimal Classification Tree with comparisons between black and non-black strike rates in each node.}
  \label{fig:tree-by-group}
\end{figure}

The question that remains is whether these criteria for exclusion are being applied equally regardless of race. To assess this, we test if there is a significant difference in strike rate between black and non-black jurors in each subgroup using Fisher's exact test~\cite{fisher1992statistical}. We adjust the p-values from the test using the Holm-Bonferroni method to account for multiple comparisons~\cite{holm1979simple}. The results of this analysis are shown in Figure~\ref{fig:tree-by-group}. We display the strike rate for black and non-black jurors that fall into each node, along with the p-value indicating whether any difference is significant. Nodes with a strike rate that is statistically significantly higher for black jurors are colored in red.

There are two cohorts where there is no statistically significant difference in strike rate between black and non-black jurors:

\begin{itemize}
    \item Node 2: potential jurors that have been accused of a crime.
    \item Node 8: potential jurors that have never been accused of a crime, do not know the defendant, do not have a family member that has been accused of a crime, but do express hesitation about imposing the death penalty
\end{itemize}

However, there are statistically significant differences in strike rate between black and non-black potential jurors in nodes 4, 6, and 9. In particular, node 4 contains jurors that have never been accused of a crime, but know the defendant. In this group, if the juror is black, the strike rate is 85\% compared to 20\% for non-black jurors. This is particularly relevant given the dissent of Justice Thomas, where he took the position that strikes by the State were appropriate because most of those dismissed knew Flowers or his family and would be biased if included in the jury. In fact, the data shows that while black jurors that knew the defendant were indeed dismissed at a very high rate, the same policy was not applied to non-black jurors that knew the defendant, who were only dismissed 20\% of the time. This clearly demonstrates a racial bias in the policy of the State: as Justice Thomas states, knowing the defendant can lead to the juror being biased, so they should be dismissed, but the State is overwhelmingly using this to exclude black jurors.

In summary, the non-linearity and interpretability of Optimal Classification Trees allowed us to identify data-driven subgroups of the juror population that exhibited similar propensities to being struck. By analyzing the strike rate for black and non-black jurors in each group, we identified two groups where the difference in strike rate was not statistically significant, but the remaining three groups did exhibit extremely significant differences based on race.

\section{Conclusions}\label{conclusions}

The original study by APM Reports used a backward stepwise logistic regression to conclude that there is significant racial disparity in the State's striking of jurors. Two of the key limitations of this approach were the heuristic nature of feature selection, and the inability to consider interactions between features.

With Optimal Feature Selection, we conducted the feature selection in an optimal fashion, and we affirmed the significance of race in the model. Our model selected all of the features used in the original model, along with four additional features that make intuitive sense. We found that race is consistently selected as one of the most predictive variables of juror strike outcome regardless of the number of features selected. This strengthens the conclusion in the original study, as we know that the feature selection of Optimal Feature Selection is exact. Furthermore, we observe a significant decrease in model performance when the race feature is removed, indicating this feature is providing unique signal in the data.

In addition, we used Optimal Classification Trees to systematically identify subgroups of the population with similar chances of being struck, and found that in two of these groups there was a significant disparity in strike rate between black and non-black jurors. These subgroups suggest systemic patterns of racial bias in the strike process, and provide direct characterizations of the situations in which black jurors are likely to have experienced discrimination. Importantly, one of the groups that exhibited significant racial disparity was jurors that knew the defendant. The dissenting opinion of Justice Thomas in the U.S. Supreme Court case argued that it was fair for the State to strike these jurors, as they would be biased on a jury, but the data shows that this reasoning was overwhelmingly used to strike black jurors rather than being applied equally to jurors of all races.

\bibliography{ref}

\begin{thebibliography}{22}
\providecommand{\natexlab}[1]{#1}
\providecommand{\url}[1]{\texttt{#1}}
\expandafter\ifx\csname urlstyle\endcsname\relax
  \providecommand{\doi}[1]{doi: #1}\else
  \providecommand{\doi}{doi: \begingroup \urlstyle{rm}\Url}\fi

\bibitem[bat(1986)]{batson}
{Batson v. Kentucky, 476 U.S. 79}, 1986.

\bibitem[fos(2016)]{foster}
{Foster v. Chatman, 578 U.S. \_\_\_}, 2016.

\bibitem[flo(2019)]{flowers}
{Flowers v. Mississippi, No. 17–9572, 588 U.S. \_\_\_}, 2019.

\bibitem[Anwar et~al.(2012)Anwar, Bayer, and Hjalmarsson]{anwar2012impact}
Shamena Anwar, Patrick Bayer, and Randi Hjalmarsson.
\newblock The impact of jury race in criminal trials.
\newblock \emph{The Quarterly Journal of Economics}, 127\penalty0 (2):\penalty0
  1017--1055, 2012.

\bibitem[Bertsimas and Copenhaver(2018)]{bertsimas2018characterization}
Dimitris Bertsimas and Martin~S Copenhaver.
\newblock Characterization of the equivalence of robustification and
  regularization in linear and matrix regression.
\newblock \emph{European Journal of Operational Research}, 270\penalty0
  (3):\penalty0 931--942, 2018.

\bibitem[Bertsimas and Dunn(2017)]{bertsimas2017optimal}
Dimitris Bertsimas and Jack Dunn.
\newblock Optimal classification trees.
\newblock \emph{Machine Learning}, 106\penalty0 (7):\penalty0 1039--1082, 2017.

\bibitem[Bertsimas and Dunn(2019)]{bertsimas2019machine}
Dimitris Bertsimas and Jack Dunn.
\newblock \emph{Machine learning under a modern optimization lens}.
\newblock Dynamic Ideas LLC, 2019.

\bibitem[Bertsimas et~al.(2016)Bertsimas, King, and
  Mazumder]{bertsimas2016best}
Dimitris Bertsimas, Angela King, and Rahul Mazumder.
\newblock Best subset selection via a modern optimization lens.
\newblock \emph{The annals of statistics}, pages 813--852, 2016.

\bibitem[Bertsimas et~al.(2017)Bertsimas, Pauphilet, and
  Van~Parys]{bertsimas2017sparse}
Dimitris Bertsimas, Jean Pauphilet, and Bart Van~Parys.
\newblock Sparse classification and phase transitions: A discrete optimization
  perspective.
\newblock \emph{arXiv preprint arXiv:1710.01352}, 2017.

\bibitem[Bertsimas et~al.(2019)Bertsimas, Pauphilet, and
  Van~Parys]{bertsimas2019sparse}
Dimitris Bertsimas, Jean Pauphilet, and Bart Van~Parys.
\newblock Sparse regression: Scalable algorithms and empirical performance.
\newblock \emph{arXiv preprint arXiv:1902.06547}, 2019.

\bibitem[Bertsimas et~al.(2020)Bertsimas, Van~Parys,
  et~al.]{bertsimas2020sparse}
Dimitris Bertsimas, Bart Van~Parys, et~al.
\newblock Sparse high-dimensional regression: Exact scalable algorithms and
  phase transitions.
\newblock \emph{The Annals of Statistics}, 48\penalty0 (1):\penalty0 300--323,
  2020.

\bibitem[Breiman et~al.(1984)Breiman, Friedman, Stone, and
  Olshen]{breiman1984classification}
Leo Breiman, Jerome Friedman, Charles~J Stone, and Richard~A Olshen.
\newblock \emph{Classification and regression trees}.
\newblock CRC press, 1984.

\bibitem[Buolamwini and Gebru(2018)]{buolamwini2018gender}
Joy Buolamwini and Timnit Gebru.
\newblock Gender shades: Intersectional accuracy disparities in commercial
  gender classification.
\newblock In \emph{Conference on fairness, accountability and transparency},
  pages 77--91, 2018.

\bibitem[Craft(2018{\natexlab{a}})]{apmreports}
Will Craft.
\newblock {APM Reports Jury Data}, 2018{\natexlab{a}}.
\newblock URL \url{https://github.com/APM-Reports/jury-data}.

\bibitem[Craft(2018{\natexlab{b}})]{apmwhitepaper}
Will Craft.
\newblock Peremptory strikes in mississippi’s fifth circuit court district,
  2018{\natexlab{b}}.
\newblock URL
  \url{https://features.apmreports.org/files/peremptory_strike_methodology.pdf}.

\bibitem[Fisher(1992)]{fisher1992statistical}
Ronald~Aylmer Fisher.
\newblock Statistical methods for research workers.
\newblock In \emph{Breakthroughs in statistics}, pages 66--70. Springer, 1992.

\bibitem[Hastie et~al.(2009)Hastie, Tibshirani, and
  Friedman]{hastie2009elements}
Trevor Hastie, Robert Tibshirani, and Jerome Friedman.
\newblock \emph{The elements of statistical learning: data mining, inference,
  and prediction}.
\newblock Springer Science \& Business Media, 2009.

\bibitem[Holm(1979)]{holm1979simple}
Sture Holm.
\newblock A simple sequentially rejective multiple test procedure.
\newblock \emph{Scandinavian journal of statistics}, pages 65--70, 1979.

\bibitem[Mehrabi et~al.(2019)Mehrabi, Morstatter, Saxena, Lerman, and
  Galstyan]{mehrabi2019survey}
Ninareh Mehrabi, Fred Morstatter, Nripsuta Saxena, Kristina Lerman, and Aram
  Galstyan.
\newblock A survey on bias and fairness in machine learning.
\newblock \emph{arXiv preprint arXiv:1908.09635}, 2019.

\bibitem[Obermeyer et~al.(2019)Obermeyer, Powers, Vogeli, and
  Mullainathan]{obermeyer2019dissecting}
Ziad Obermeyer, Brian Powers, Christine Vogeli, and Sendhil Mullainathan.
\newblock Dissecting racial bias in an algorithm used to manage the health of
  populations.
\newblock \emph{Science}, 366\penalty0 (6464):\penalty0 447--453, 2019.

\bibitem[Tibshirani(1996)]{tibshirani1996regression}
Robert Tibshirani.
\newblock Regression shrinkage and selection via the lasso.
\newblock \emph{Journal of the Royal Statistical Society: Series B
  (Methodological)}, 58\penalty0 (1):\penalty0 267--288, 1996.

\bibitem[Zou and Hastie(2005)]{zou2005regularization}
Hui Zou and Trevor Hastie.
\newblock Regularization and variable selection via the elastic net.
\newblock \emph{Journal of the royal statistical society: series B (statistical
  methodology)}, 67\penalty0 (2):\penalty0 301--320, 2005.

\end{thebibliography}

\end{document}